\documentclass[conference]{IEEEtran}

\usepackage{amssymb}
\usepackage{graphicx}
\usepackage{amsmath}
\usepackage{geometry}
\usepackage{booktabs}
\usepackage{siunitx}
\usepackage{float}
\usepackage{subcaption}
\usepackage{hyperref}
\usepackage{bm}

\begin{document}
\title{Keypoint Detection Technique for Image-Based Visual Servoing of Manipulators}

\author{\IEEEauthorblockN{Niloufar Amiri}
    \IEEEauthorblockA{Department of Mechanical,\\
    Industrial, and Mechatronics Engineering\\
    Toronto Metropolitan University\\
    Toronto, Canada\\
    Email: niloufar.amiri@torontomu.ca}
    \and
    \IEEEauthorblockN{Guanghui Wang}
    \IEEEauthorblockA{Department of Computer Science\\
    Toronto Metropolitan University\\
    Toronto, Canada\\
    Email: wangcs@torontomu.ca}
    \and
    \IEEEauthorblockN{Farrokh Janabi-Sharifi}
    \IEEEauthorblockA{Department of Mechanical,\\
    Industrial, and Mechatronics Engineering\\
    Toronto Metropolitan University\\
    Toronto, Canada\\
    Email: fsharifi@torontomu.ca}
}
\newgeometry{top=1in,bottom=0.75in,right=0.75in,left=0.75in}
\maketitle

\begin{abstract}

This paper introduces an innovative keypoint detection technique based on Convolutional Neural Networks (CNNs) to enhance the performance of existing Deep Visual Servoing (DVS) models. To validate the convergence of the Image-Based Visual Servoing (IBVS) algorithm, real-world experiments utilizing fiducial markers for feature detection are conducted before designing the CNN-based feature detector. To address the limitations of fiducial markers, the novel feature detector focuses on extracting keypoints that represent the corners of a more realistic object compared to fiducial markers. A dataset is generated from sample data captured by the camera mounted on the robot end-effector while the robot operates randomly in the task space. The samples are automatically labeled, and the dataset size is increased by flipping and rotation. The CNN model is developed by modifying the VGG-19 pre-trained on the ImageNet dataset. While the weights in the base model remain fixed, the fully connected layer's weights are updated to minimize the mean absolute error, defined based on the deviation of predictions from the real pixel coordinates of the corners. The model undergoes two modifications: replacing max-pooling with average-pooling in the base model and implementing an adaptive learning rate that decreases during epochs. These changes lead to a 50 percent reduction in validation loss. Finally, the trained model's reliability is assessed through k-fold cross-validation.
\end{abstract}


\IEEEpeerreviewmaketitle

\section{Introduction}

Visual servoing (VS) is a popular control method used to enable robotic systems, including fixed or mobile manipulators with high degrees of freedom, to operate in unstructured environments \cite{abdulhafiz2022deep, machkour2022classical}. This method relies on sensory feedback from one or more cameras within the control loop. The information, typically referred to as features, is supplied by the camera to estimate the current configuration of the robot and guide the system toward the desired configuration. The camera can be mounted on the robot's end-effector in an eye-in-hand configuration \cite{kelly2000stable, muthusamy2021neuromorphic, chang2018robotic, wu2019unsupervised}. Alternatively, it can be fixed in a position in the world space, known as an eye-to-hand configuration \cite{tokuda2021convolutional}, or a combination of these configurations \cite{lippiello2005eye}. 

Position-based visual servoing (PBVS) and image-based visual servoing (IBVS) represent two fundamental approaches among various visual servoing techniques \cite{janabi2010comparison}. In PBVS, estimating the relative position and orientation (pose) of the target object with respect to the camera depends on image features and camera calibration. This makes PBVS highly sensitive to precise object geometry measurement and accurate camera calibration \cite{zhou2022position}. In contrast, IBVS implicitly defines the goal pose based on the image of the target at its desired position. As features reach their desired values in the image plane, the camera (and the robot) achieves its desired pose relative to the object in the world frame. This simplifies the pose control problem to pixel coordinate control in 2D image space. Accordingly, IBVS has gained popularity for its robustness against inaccuracies in target modeling and camera calibration errors \cite{fioravanti2008image,liang2021fully}. The fact that IBVS relies entirely on image features highlights the significance of feature detection as a crucial research area in this field.

The focus of this study is on detecting interest points, also known as keypoints, as an integral part of IBVS. This process relies on identifying specific features and their precise pixel coordinates in the image. The proposed detection algorithm also has the potential to be applied in other applications involving the detection of specific features in an image. To accomplish the feature detection task, there are several approaches. One approach involves defining keypoints as local extrema in a 3-dimensional space scale. This enables the application of classic feature detection methods, including the partial differentiation of the Laplacian \cite{mikolajczyk2005comparison, mukherjee2015comparative} and non-linear partial differential equations \cite{alcantarilla2012kaze, salti2013keypoints}. However, these conventional methods lack the intelligence to differentiate between features and select among them. The reason is that the selection typically relies on examining a large neighborhood of features in the image plane, which is not possible with conventional methods. This limitation renders conventional methods unsuitable for IBVS, which emphasizes detecting and tracking very precise keypoints among numerous other features. To address this issue, other classic methods, such as color filters and Hough transforms, can be employed. Nevertheless, these methods are computationally expensive and sensitive to noise and illumination changes.
\newgeometry{top=0.75in,bottom=0.75in,right=0.75in,left=0.75in}
Another approach involves using fiducial markers like AprilTags, QR codes, and ArUco for extracting key features through template matching. This method facilitates the robust and precise extraction of interest features with minimal computational effort. However, it faces two significant constraints. Firstly, these tags lack a visually appealing industrial appearance \cite{jurado2021design}. More importantly, they impose a constraint on feature detection by altering the scene. As a result, the algorithm is unable to detect specific features of an arbitrary object. This constraint is relevant in IBVS as the ultimate objective is to enable the robot's operation in unstructured environments \cite{li2023hybrid}. The limitations of conventional techniques motivated us to adopt AI-based methods.

The supervised CNN-based approaches offer a more effective way of feature detection, surpassing the limitations of previous strategies \cite{ma2022semantic,al2022robotic, he2022deep}. These methods are versatile and applicable in various real-world scenarios due to their generalization and lack of reliance on templates. In CNN-based feature detection, the coordinates of key features are fed into the model as image labels. This simplifies the feature detection to a regression task, where the inputs are RGB images captured by the camera, and the outputs are the pixel coordinates of keypoints \cite{lee2017learning}.
\begin{figure}[t]
    \centering
    \includegraphics[width = 7 cm]{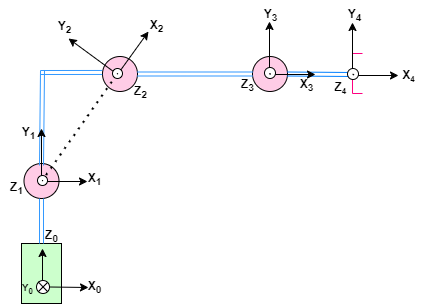}
    \caption{Schematic of joints and links}
    \label{jointlink}
\end{figure}

This paper offers three main contributions. Firstly, it introduces an innovative feature detection algorithm based on deep learning. This algorithm modifies the VGG-19 deep neural network \cite{mascarenhas2021comparison, rajinikanth2020customized, bansal2021transfer} through transfer learning to tailor it for IBVS applications.  The simpler yet deeper structure of VGG-19, with its several convolutional layers, makes it more suitable for extracting delicate features such as keypoints compared to other networks \cite{xiao2020application, shah2023comparing}. The model is then trained using an automatically generated dataset that includes images of a target object in various configurations of the robot. Secondly, despite previous DVS models \cite{machkour2022classical, bansal2021transfer, al2022robotic}, the trained model presented in this paper is independent of the reference pose of the camera. Thirdly, the paper presents a derived kinematic model of the openMANIPULATOR-X and introduces an IBVS control algorithm for manipulating the end-effector's pose. The IBVS algorithm's experimental validation is also conducted using AprilTags for feature detection. 

The remainder of the paper is organized as follows: In Section II, the theory of IBVS is presented, followed by the analysis of the results obtained from implementing the algorithm on the OpenMANIPULATOR-X. Section III covers the architecture of the CNN, along with the description of dataset generation, model training, testing results, and k-fold cross-validation. Finally, the highlights of the paper are discussed in Section IV.

\section{Image-Based Visual Servoing}
\subsection{OpenMANIPULATOR Kinematics}

\begin{figure}[t]
    \centering
    \includegraphics[scale=0.1]{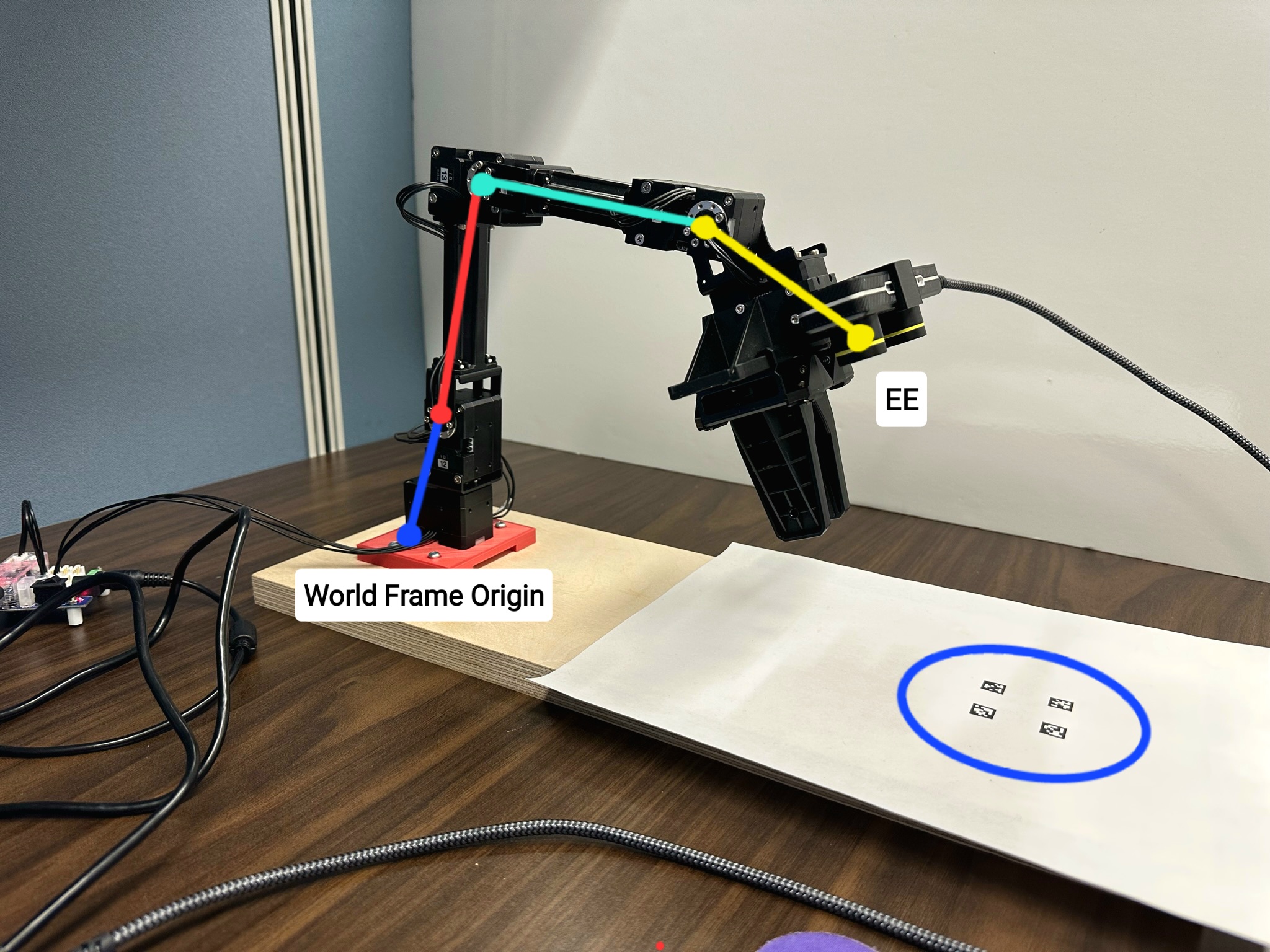}
    \caption{The mounted camera on the robot with an eye-in-hand configuration}
    \label{real}
\end{figure}

Fig. \ref{jointlink} shows a schematic of the OpenMANIPULATOR-X skeleton and the attached coordinate system. The robot has four links and four revolute joints, with the last three joints being parallel. The camera is mounted on the tip of the manipulator, as shown in Fig. \ref{real}, and AprilTags are placed on a horizontal plane in front of the robot.
Table I shows the DH parameters for this configuration.

\begin{table}[t]
    \centering
    \caption{DH Parameters of OpenMANIPULATOR-X}
    \begin{tabular}{|c|c|c|c|c|c|}
        \hline
        Link & $a_i (m)$ & $\alpha_i (m)$ & $d_i(m)$ & $\theta_i (rad)$ & $\theta_{i_0} (rad)$\\
        \hline
        1 & 0 & $\frac{\pi}{2}$ & 0.077 & $\theta_1$&0\\
        \hline
        2 & 0.13 & 0 & 0 & $\theta_2$& $\mathrm{atan(128/24)}$\\
        \hline
        3 & 0.124 & 0 & 0 & $\theta_3$&$\mathrm{-atan(128/24)}$\\
        \hline
        4 & 0.126 & 0 & 0 & $\theta_4$&0\\
        \hline
    \end{tabular}
\end{table}

\subsection{IBVS Theory}
The IBVS control law, which establishes the relationship between the camera velocity and the errors in image features, can be expressed as \cite{corke2023robotics}

\begin{equation}
    {\bm{V}} = \mathit{\lambda} \begin{bmatrix}
        {\bm{J}_p}(\bm{p}_1, \mathit{Z}_1) \\
        \vdots \\
        {\bm{J}_p}(\bm{p}_N, \mathit{Z}_N)
    \end{bmatrix}^+ (\bm{p}^* - \bm{p}),
    \label{eq1}
\end{equation}

\noindent where $\bm{V} = (v_x, v_y, v_z, \omega_x, \omega_y, \omega_z) \in \mathbb{R}^6$ is the vector of body angular and translational velocity of the camera. $\lambda \in \mathbb{R}$ is the control gain. $\bm{J}_p$ is the $2\times 6$ interaction matrix or image Jacobian. $N \in \mathbb{Z}$ represents the index of features. $Z \in \mathbb{R}$ is the estimate of camera depth for each feature. $\bm{p}^*$ and $\bm{p}$ are $1\times 2$ vectors denoting the desired and current pixel coordinates of features.

Once the camera velocity is obtained, the angular velocity of the joints is updated through the inverse kinematics of the manipulator

\begin{equation}
\dot{\boldsymbol{q}}=\boldsymbol{J}^{-1}(\boldsymbol{q})\bm{V},
\end{equation}

\noindent where $\bm{q}$ and $\bm{\dot{q}}$ represent the angular position and velocity of joints, respectively. $\boldsymbol{J}$ is the $6\times 4$ Jacobian for the camera velocity in the camera frame, which can be obtained from the Jacobian for the robot end-effector in the world frame ($\boldsymbol{J}_{a}$) according to the following mapping:

\begin{equation}
\boldsymbol{J} = \begin{bmatrix}
R_c & 0 \\
0 & R_c
\end{bmatrix} \boldsymbol{J}_{a},
\end{equation}

\noindent where $R_c$ is the rotation of the camera in the world frame.

One of the challenging issues of the algorithm is computing the feature error corresponding to the difference between the location of specific pixels in an image relative to their desired location. The control law guarantees the convergence of current pixel coordinates to their desired pixel coordinates while the control parameter ($\lambda$) is selected properly.

\begin{figure}[t]
    \centering
    \includegraphics[width = 8 cm]{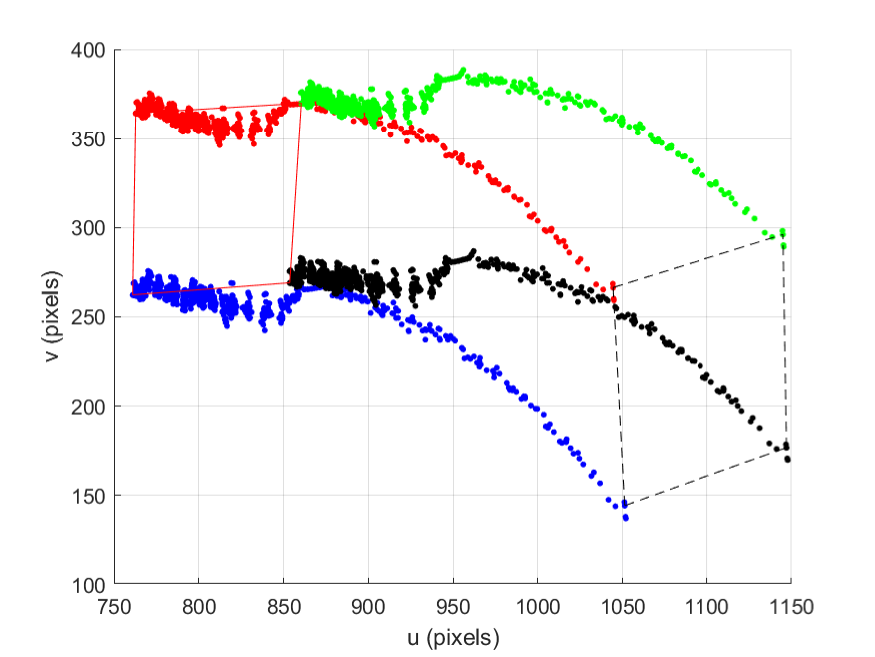}
    \caption{The change in pixel coordinates of the Keypoints over time, from the initial position (black quadrilateral) to the desired position (red quadrilateral)}
    \label{fig:coordinates}
\end{figure}

\subsection{IBVS Validation}
Fig. \ref{fig:coordinates} visualizes the change in pixel coordinates of four keypoints over time by implementing the IBVS algorithm along with the inverse kinematics of OpenMANIPULATOR-X in a real experiment. The control parameter $\lambda$ is set to 1 and the time step is 0.005 for 1500 iterations (7.5s). The results show that the pixel coordinates converge to their desired values over time. In each iteration, a snapshot is captured from the scene, and the centers of four AprilTags are obtained. The desired camera velocity is then determined based on the computed feature error. Since the location of the camera is considered as the end-effector position, the problem is simplified to an inverse kinematics problem. The numerical inverse kinematic algorithm is then employed to find the desired joint velocity from the end effector (camera) velocity.

\begin{figure}[t]
    \centering
    \includegraphics[width = 8 cm]{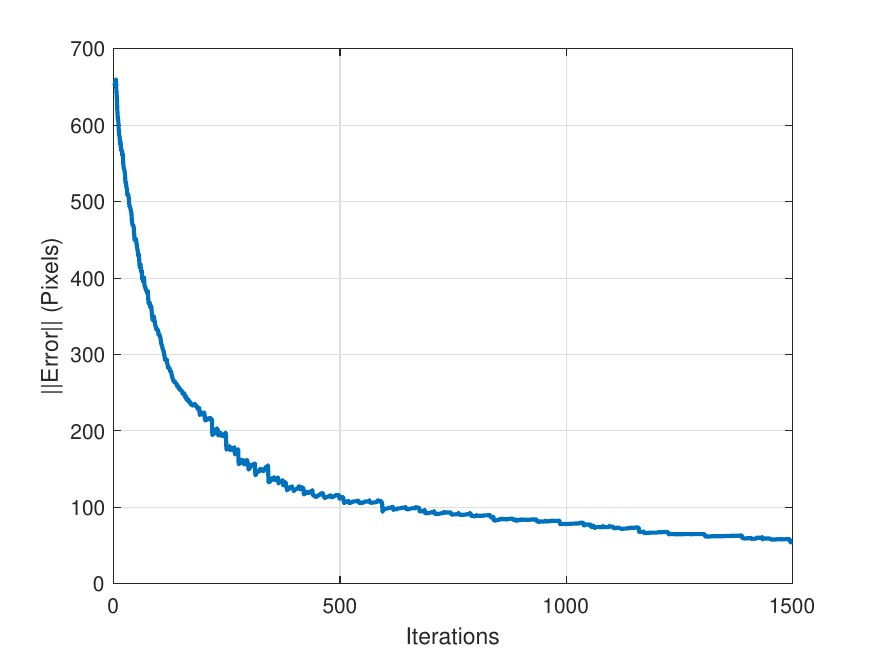}
    \caption{Second norm of the pixel coordinate error for four keypoints}
    \label{fig:Error}
\end{figure}

Fig. \ref{fig:Error} shows the second norm of pixel coordinates error over time. Overall, the error is decreasing as the robot gets closer to its desired configuration.

To improve the performance of this algorithm for more realistic scenarios, a CNN-based feature detector is designed that is thoroughly investigated in the next section.

\section{Keypoint Feature Detection}

\subsection{Dataset Generation}
\begin{figure*}[t]
    \centering
    
    \begin{subfigure}{0.32\linewidth}
        \centering
        \includegraphics[width=\linewidth]{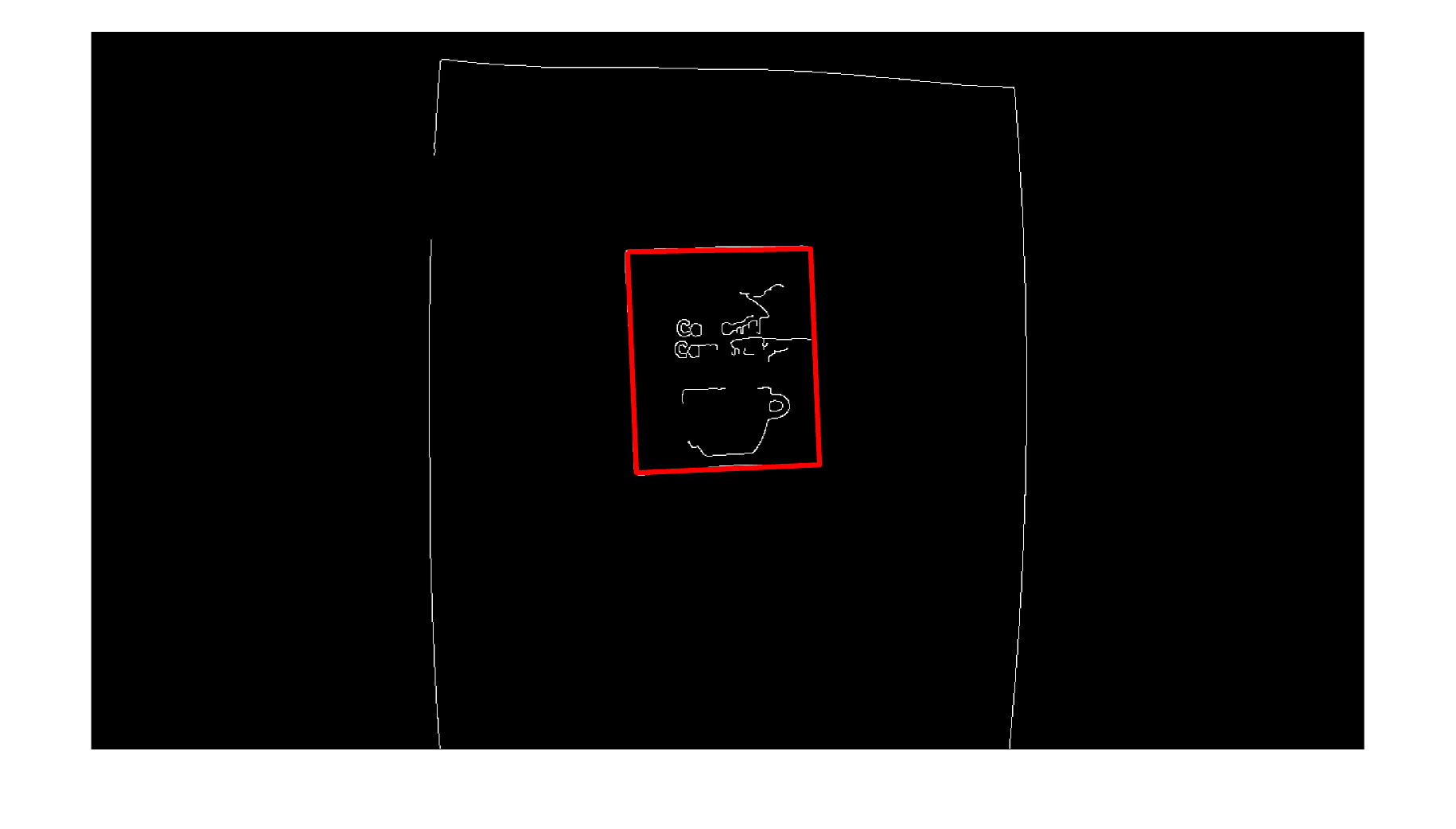}

    \end{subfigure}
    \hfill 
    \begin{subfigure}{0.32\linewidth}
        \centering
        \includegraphics[width=\linewidth]{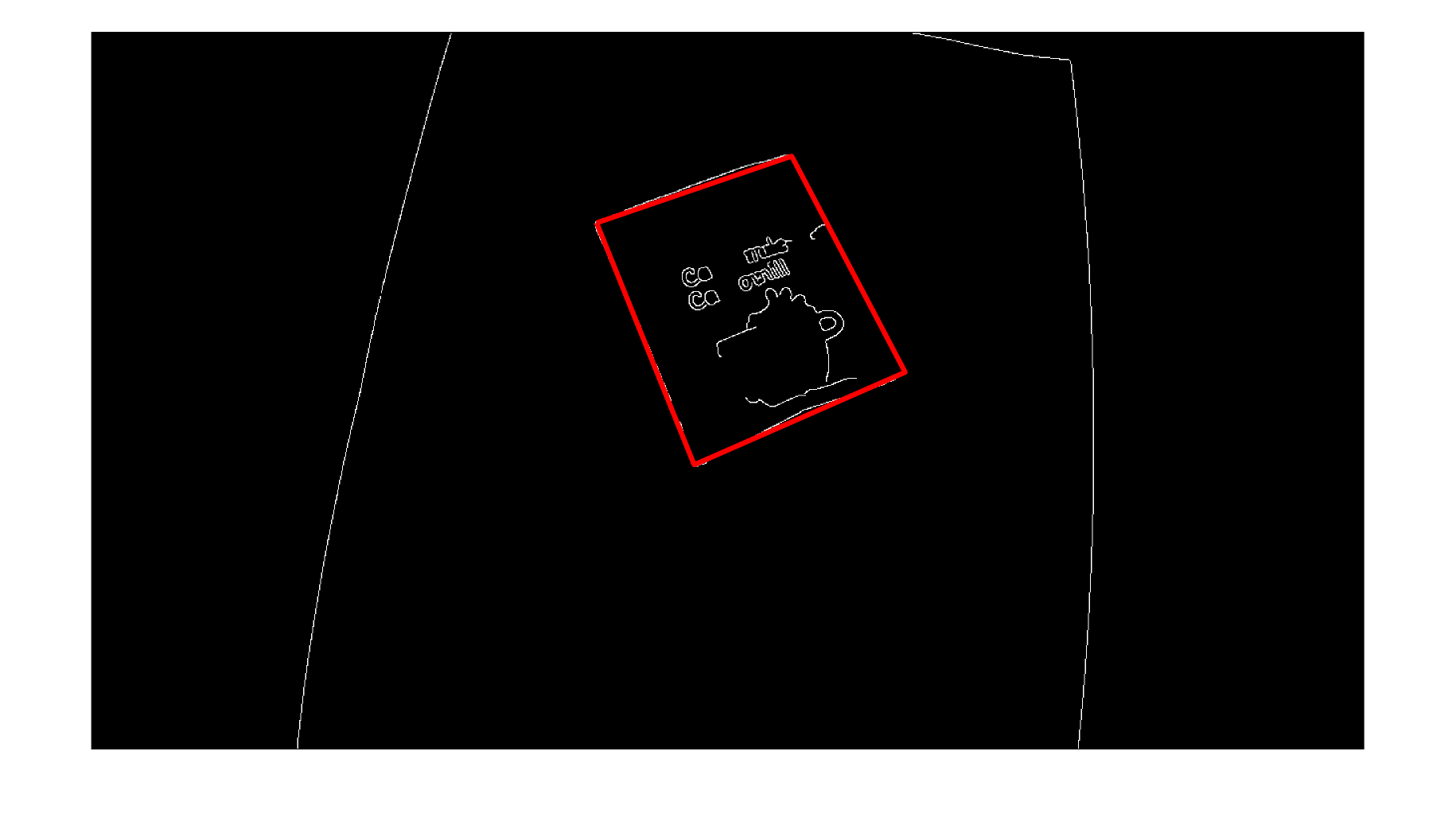}

    \end{subfigure}
    \hfill 
    \begin{subfigure}{0.32\linewidth}
        \centering
        \includegraphics[width=\linewidth]{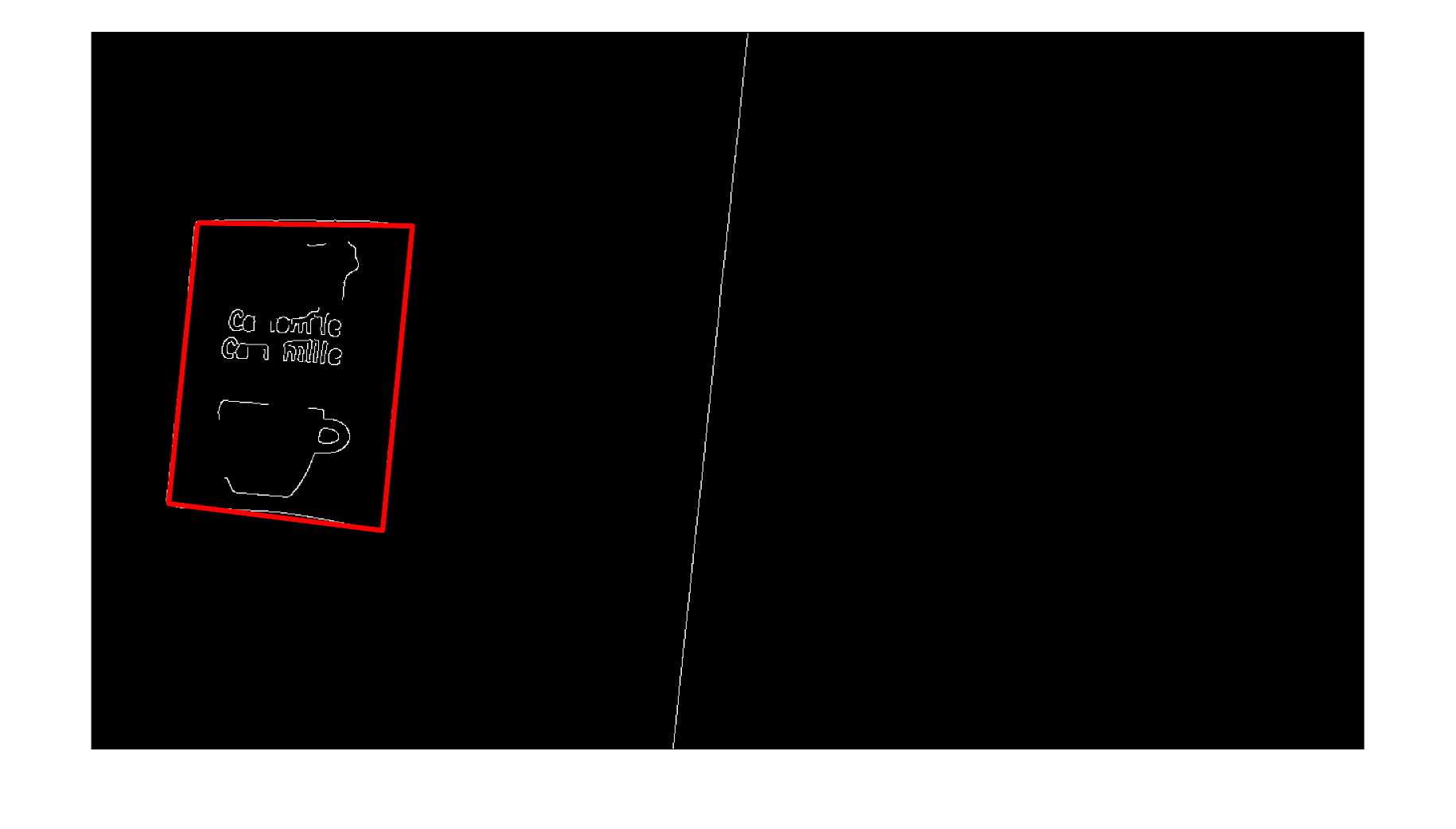}

    \end{subfigure}
    
    \caption{Examples of annotated images created by conventional image processing techniques}
    \label{fig:annotate}
\end{figure*}
\begin{figure*}[t]
    \centering
    
    \begin{subfigure}{0.28\linewidth}
        \centering
        \includegraphics[width=\linewidth]{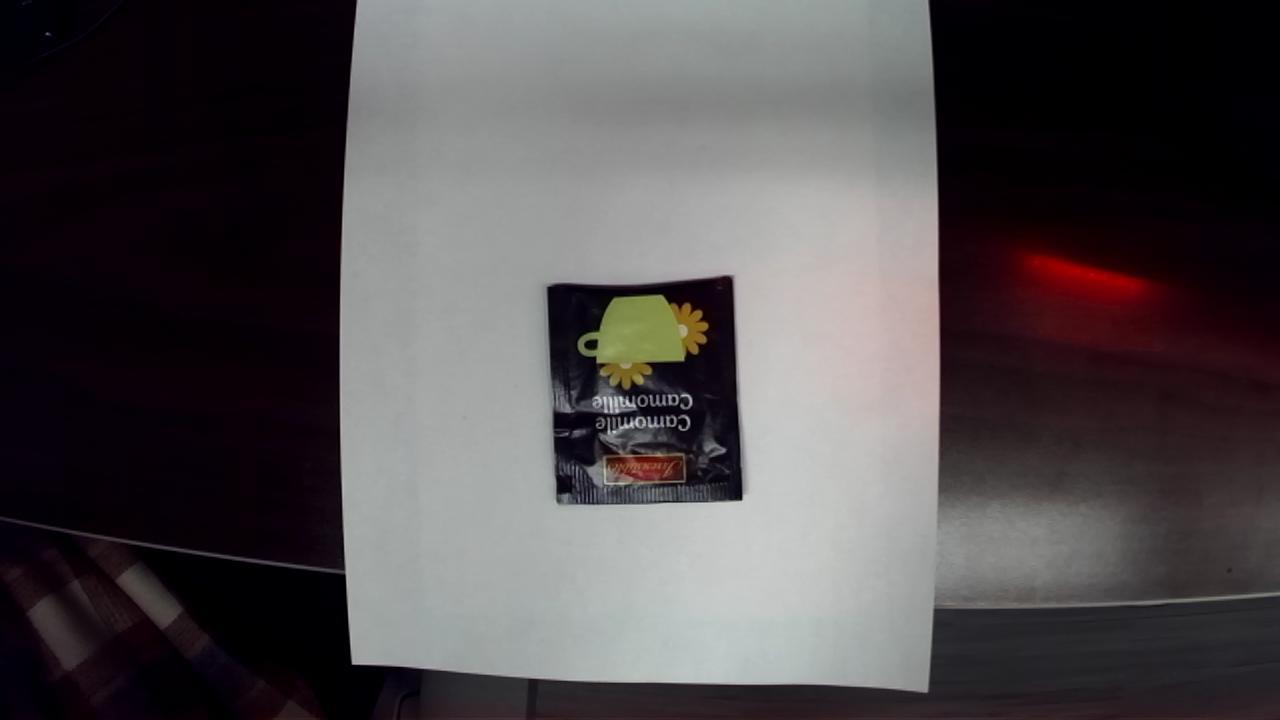}
    \end{subfigure}
    \hfill 
    \begin{subfigure}{0.28\linewidth}
        \centering
        \includegraphics[width=\linewidth]{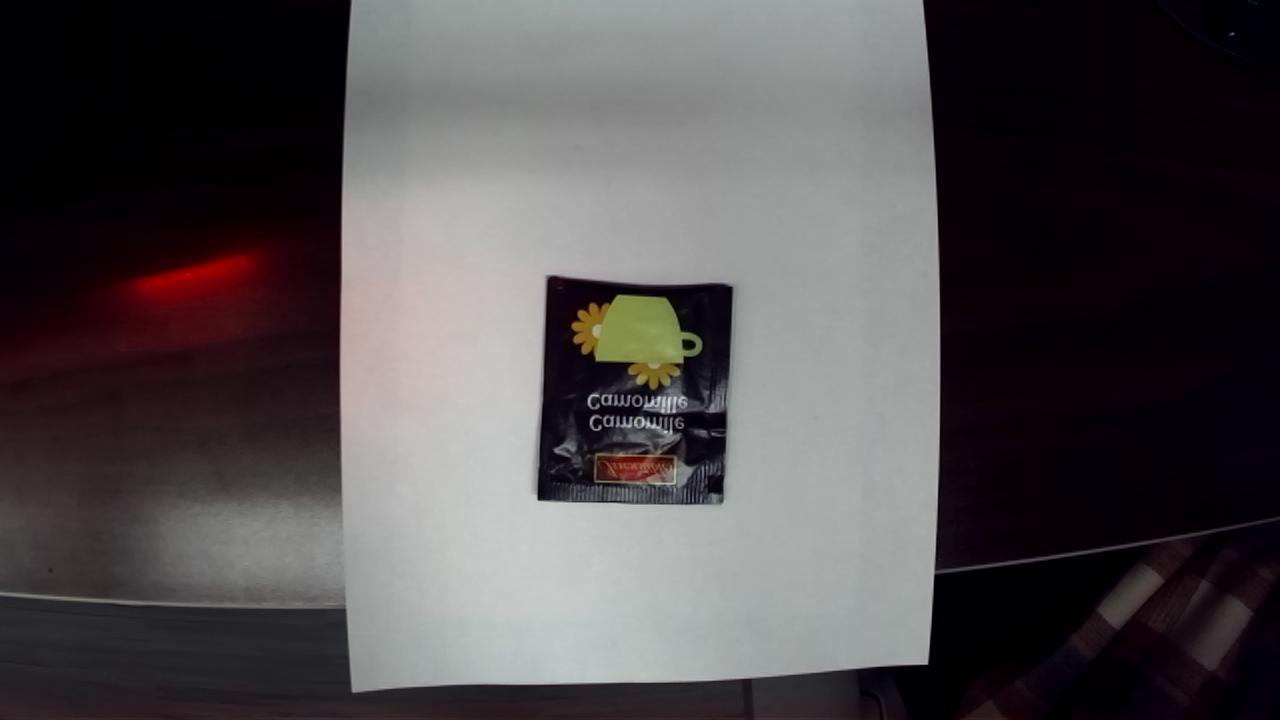}
    \end{subfigure}
    \hfill 
    \begin{subfigure}{0.28\linewidth}
        \centering
        \includegraphics[width=\linewidth]{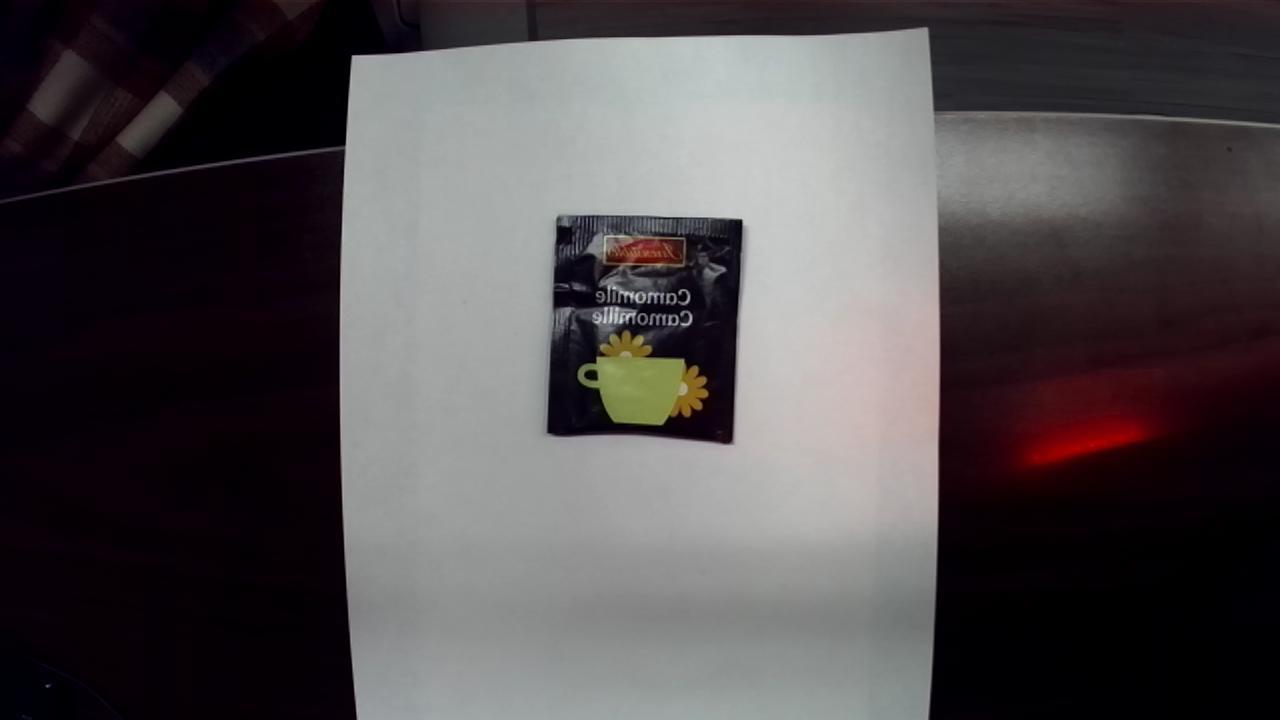}
    \end{subfigure}
    
    \caption{Examples of images created by rotating and flipping the original images}
    \label{fig:augment}
\end{figure*}

The first step before training a CNN model is to create the dataset. The target object in this case is a tea bag with four corners. The tea bag is positioned in front of the robot within its field of view. Subsequently, 400 pictures were captured by the robot's camera, each depicting the tea bag in various positions. The images were captured by the left camera of the ZED mini (StereoLabs, Paris, France) stereo camera. The camera was calibrated before taking the images. The calibration parameters using algorithm \cite{fetic2012procedure} are listed in Table II.

The images were labeled based on the four corners of the target object, with careful consideration given to the order of these corners. The top-left corner was considered the first label, and the bottom-left corner was the fourth label. A combination of image processing techniques was used to automatically annotate the images. The annotation process includes converting the images into grayscale and then applying a Canny edge detector with a specific threshold. Finally, the quadrilateral is detected by finding the maximum and minimum intersections of edges with parallel lines having a certain slope. Some of the annotated samples are shown in Fig. \ref{fig:annotate}. After partitioning the dataset into training and testing categories, it was downloaded as a zip file. The labels were stored in xlsx files containing the coordinates of the corners in a specific order. Further processing, including data augmentation, was performed subsequently.

\begin{table}[t]
\label{table1}
    \centering
    \caption{ZED mini Camera Parameters}
    \begin{tabular}{|c|c|}
        \hline
        $u_0$ (pixels) & 617.930 \\
        \hline
        $v_0$ (pixels) & 366.566 \\
        \hline
        $\rho $ (m) & $4 \times 10^{-6}$ \\
        \hline
        $f$ (pixels) & $[686.015, 681.838]$ \\
        \hline
    \end{tabular}
\end{table}

\subsection{Data Augmentation}

The use of data augmentation techniques was imperative in addressing the challenges posed by the relatively small-sized dataset. However, an issue arose with automatic data augmentation, where changes in the position of corners during augmentation did not correspondingly update the labels due to the absence of a mathematical relationship between images and labels. As an alternative approach, augmented images were generated through programming. This involved rotating or flipping the image and then updating the labels accordingly based on the performed transformation. Three types of image augmentation were implemented, including 180-degree rotation, as well as both vertical and horizontal flipping as shown in Fig. \ref{fig:augment}. This process effectively expanded the dataset size by a factor of 4.

\subsection{Convolutional Neural Network Architecture}
The CNN model consists of two sub-models. The base model, VGG-19, comprises five blocks summarized in Table 2. Each block includes two or more convolutional layers for feature extraction and a max-pooling layer to retain the most significant features. However, in the modified version, the max-pooling layer is replaced by average-pooling to overcome overfitting. The outputs are flattened and passed through a fully connected layer to facilitate the regression task and predict eight coordinates.

\begin{table}[t]
\label{table2}
    \centering
    \caption{Modified VGG-19 CNN Architecture}
    \begin{tabular}{@{}lccc@{}}
        \toprule
        \textbf{Block Name} & \textbf{Input Layer} & \textbf{Output Layer} \\
        \midrule
        \textbf{Block 1} & & \\
        Convolution 2D & $[180,320,3]$ & $[180,320,64]$ \\
        Convolution 2D & $[180,320,64]$ & $[180,320,64]$ \\
        Average Pooling 2D & $[180,320,64]$ & $[90,160,64]$ \\
        \midrule
        \textbf{Block 2} & & \\
        Convolution 2D & $[90,160,64]$ & $[90,160,128]$ \\
        Convolution 2D & $[90,160,128]$ & $[90,160,128]$ \\
        Average Pooling 2D & $[90,160,128]$ & $[45,80,128]$ \\
        \midrule
        \textbf{Block 3} & & \\
        Convolution 2D & $[45,80,128]$ & $[45,80,256]$ \\
        Convolution 2D & $[45,80,256]$ & $[45,80,256]$ \\
        Convolution 2D & $[45,80,256]$ & $[45,80,256]$ \\
        Convolution 2D & $[45,80,256]$ & $[45,80,256]$ \\
        Average Pooling 2D & $[45,80,256]$ & $[22,40,256]$ \\
        \midrule
        \textbf{Block 4} & & \\
        Convolution 2D & $[22,40,256]$ & $[22,40,512]$ \\
        Convolution 2D & $[22,40,512]$ & $[22,40,512]$ \\
        Convolution 2D & $[22,40,512]$ & $[22,40,512]$ \\
        Convolution 2D & $[22,40,512]$ & $[22,40,512]$ \\
        Average Pooling 2D & $[22,40,512]$ & $[11,20,512]$ \\
        \midrule
        \textbf{Block 5} & & \\
        Convolution 2D & $[11,20,512]$ & $[11,20,512]$ \\
        Convolution 2D & $[11,20,512]$ & $[11,20,512]$ \\
        Convolution 2D & $[11,20,512]$ & $[11,20,512]$ \\
        Convolution 2D & $[11,20,512]$ & $[11,20,512]$ \\
        Average Pooling 2D & $[11,20,512]$ & $[5,10,512]$ \\
        \midrule
        Flatten & $[5,10,512]$ & $[25600,1]$ \\
        \midrule
        Dense & $[25600,1]$ & $[8,1]$ \\
        \bottomrule    
    \end{tabular}
    
\end{table}

\subsection{Training}

In this paper, transfer learning techniques were employed in the Tensorflow framework to take advantage of pre-trained deep CNNs capable of feature detection. Specifically, we utilized a VGG-19 model pre-trained on the ImageNet dataset. During training, we froze the weights of the VGG-19 convolutional layers while adapting the fully connected dense layer with a linear activation function. This adaptation aimed to minimize the selected loss function, which, in this case, is the Mean Absolute Error (MAE). The Adam optimizer is selected to train the deep CNN in 300 epochs with a batch size of 16 and an initial learning rate of $10^{-5}$. The data split for training and validation is set to 0.1. 

\begin{figure} [t]
    \centering
    \includegraphics[width = 7 cm]
    {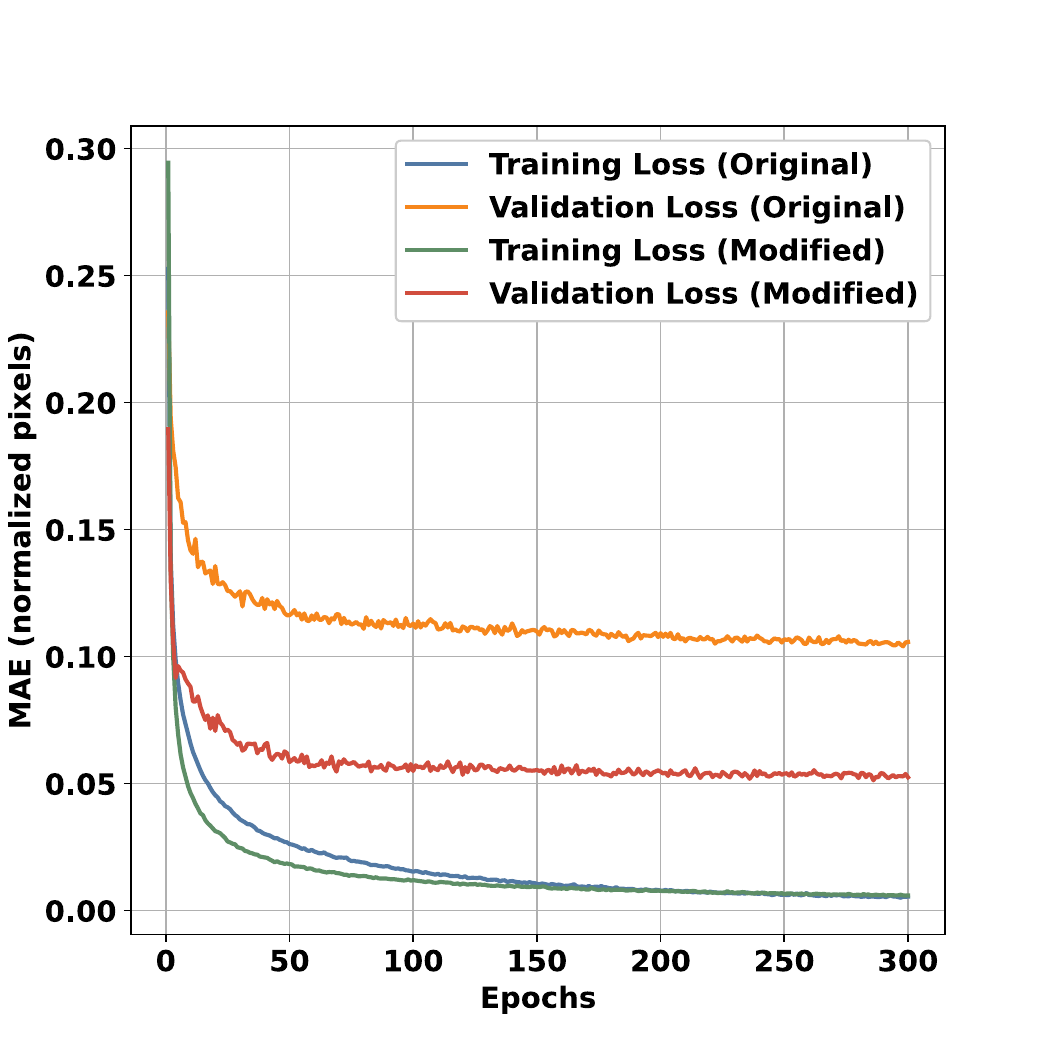}
    \caption{Learning Curve: training and validation loss updates in the original and modified models}
    \label{Learning Curve}
\end{figure}

The learning curve concerning the original model without average pooling is shown in Fig. \ref{Learning Curve}. From the plot, we can see that the training loss dropped from approximately 0.3 to about 0.007 pixels for normalized pixel values. The validation curve outlines how the validation loss has evolved with the epochs. Likewise, the validation loss dropped from 0.25 to roughly above 0.1, showing that the model is experiencing over-fitting. To further enhance the training results, the modified network is re-trained for a second time with average-pooling instead of max-pooling and an adaptive learning rate starting from $10^{-5}$ and reducing every 2500 steps by a factor of 0.95 percent. The result of the second training is illustrated in Fig. \ref{Learning Curve}. The same pattern is observable for the training dataset. Similar to the first model, the training loss decreased from just below 0.3 to 0.0055. However, the result shows a significant improvement in the validation loss which is approximately half of the one in the original model. 

\subsection{Testing}

\begin{figure}[t]
    \centering
    \includegraphics[width = 7 cm]
    {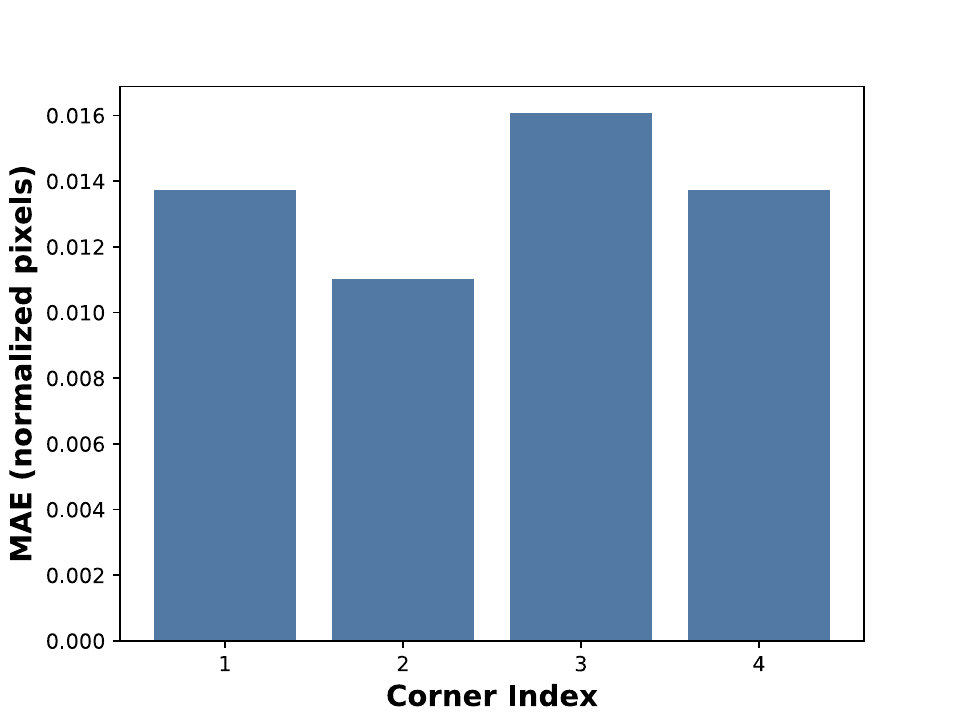}
    \caption{The mean absolute error between each predicted corner value and the actual label value}
    \label{corner-err}
\end{figure}

\begin{figure*}[t]
    \centering
    
    \begin{subfigure}{0.3\linewidth}
        \centering
        \includegraphics[width=\linewidth]{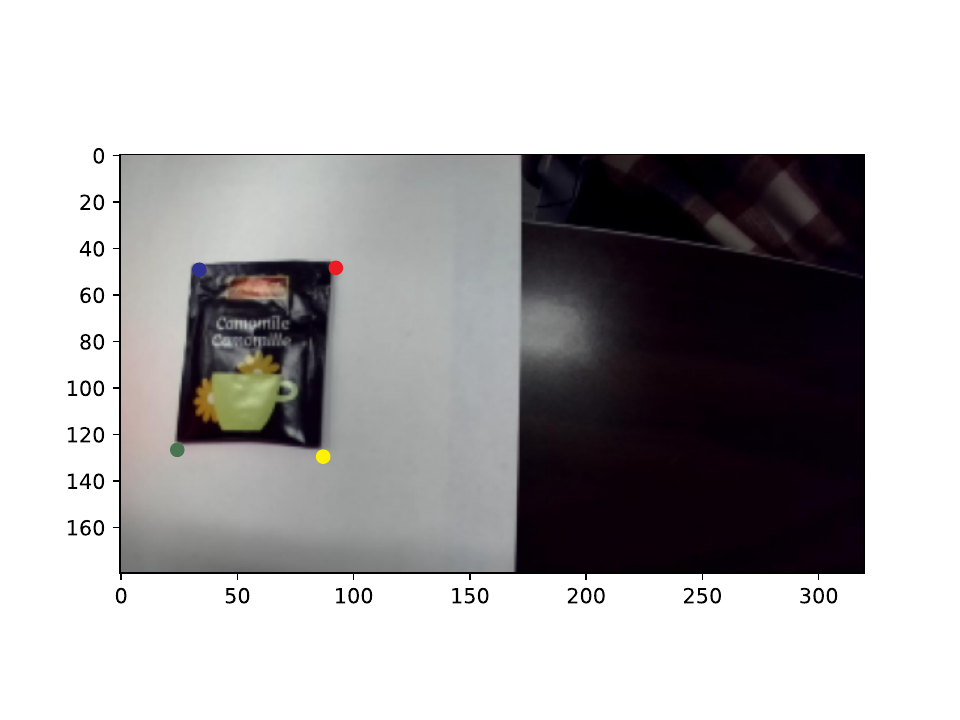}

    \end{subfigure}
    \hfill 
    \begin{subfigure}{0.3\linewidth}
        \centering
        \includegraphics[width=\linewidth]{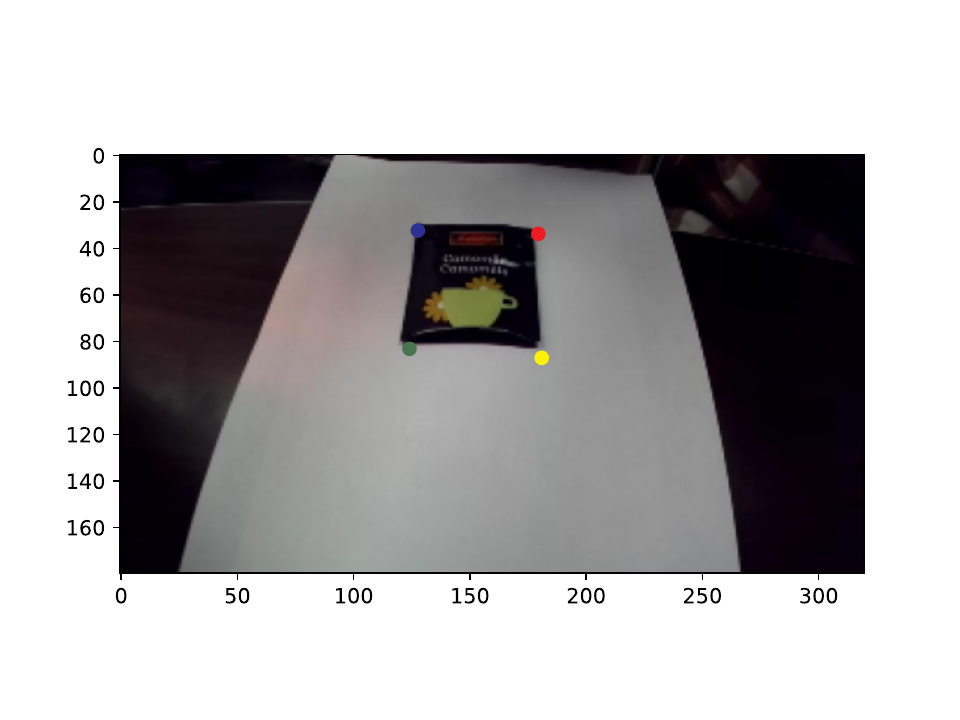}

    \end{subfigure}
    \hfill 
    \begin{subfigure}{0.3\linewidth}
        \centering
        \includegraphics[width=\linewidth]{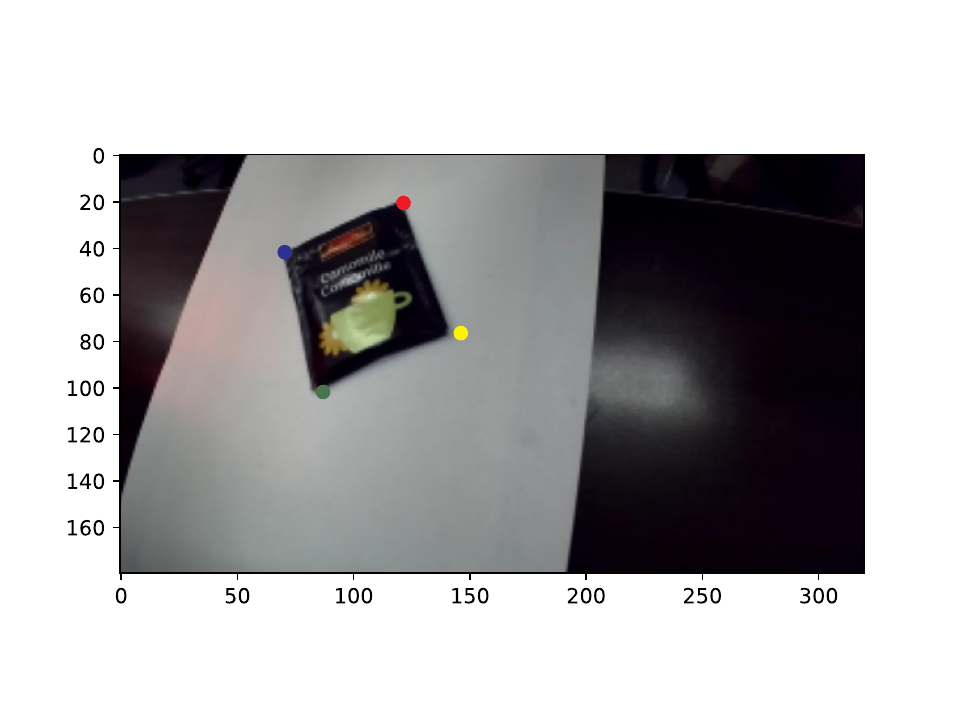}

    \end{subfigure}
    
    \caption{Examples of feature detection accurately identifying corners in unseen images, with corner indices highlighted in different colors.}
    \label{fig:test}
\end{figure*}

Forty unseen images are selected for testing the trained CNN. The mean absolute error for the deviation of each predicted corner from the real label values is shown in Fig. \ref{corner-err}. It should be noted that corner 3 (bottom right) experiences the maximum deviation, primarily attributable to the lighting conditions of the scene. A shadow in the vicinity of this corner makes it challenging to accurately locate it. As shown in Fig. \ref{corner-err}, the maximum mean absolute error in the prediction is less than 0.016 normalized pixel value, which is a relatively satisfactory result for our application. Fig. \ref{fig:test} illustrates the predicted corners for three unseen sample images. The accuracy in positioning the corners and their respective orders verifies the effectiveness of the modified model performance in feature extraction.

\subsection{K-fold Cross Validation}

\begin{figure}[t]
    \centering
    \includegraphics[width = 7 cm]
    {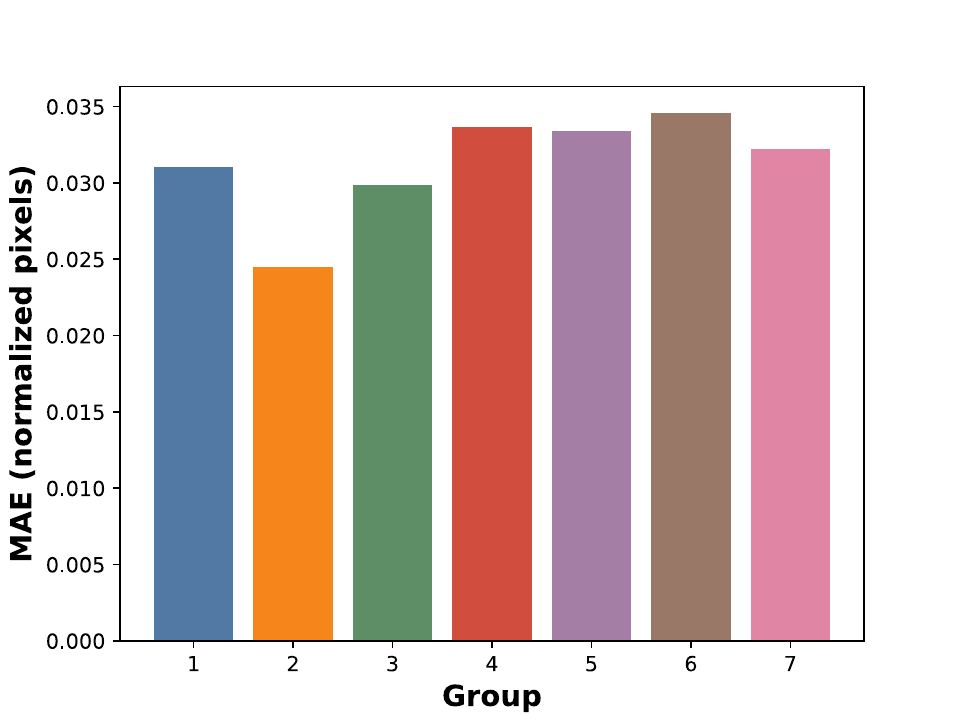}
    \caption{K-fold cross-validation results: Mean of absolute error for each test group}
    \label{cross}
\end{figure}

K-fold cross-validation is used to assess the reliability of the model by measuring the accuracy of the model for variant test images. In this procedure, the dataset is randomly partitioned into seven groups. During each training round, six groups are utilized for training, and one group is reserved for testing the model's performance. This cycle is iterated seven times until all groups have been employed for testing the model. Figure \ref{cross} illustrates that the model's accuracy remains relatively consistent across various sets of test data. 

\section{Conclusion}
This paper employs deep CNN-based techniques to address the challenge of detecting specific features in a specific order for IBVS of manipulators. Compared to other methods such as known visual markers, this approach offers a more realistic and feasible keypoint detection method, especially for operating in unstructured environments. The CNN model utilized in this study has several distinguishing features. Firstly, the method used for labeling images in a specific order reduced the required time for creating a dataset compared to manual annotation. Secondly, the data augmentation technique used in this paper differs from standard methods, in which image labels are modified according to the changes applied to the image. Thirdly, replacing the max pooling layer with the average pooling layer significantly improved the accuracy of the model on the validation dataset, which was a crucial measure since the model was prone to significant overfitting.

Future work includes generating a larger dataset and assessing the performance of the trained model in a real experiment in the presence of robot vibration and other real-world parameters.

\bibliographystyle{ieeetr}
\bibliography{ref}

\end{document}